\documentclass[10pt,conference,letterpaper]{IEEEtran}
\usepackage{amsmath,amssymb}
\usepackage{graphicx}
\usepackage{caption}
\usepackage{subcaption}
\usepackage{cite}
\usepackage{url}
\usepackage[hidelinks]{hyperref}

\graphicspath{{figs/}}

\begin{document}

\title{Walking on the Slope: Stable Bipedal Gaits with Genetic-Algorithm-Optimized Trajectories}

\author{\IEEEauthorblockN{Madhav Rijal}
\IEEEauthorblockA{Department of Mechanical Engineering, Indian Institute of Technology Kanpur, Kanpur 208016, India\\
rijalmadhav12@gmail.com}}

\maketitle

\begin{abstract}
This paper presents the kinematic and dynamic modeling, trajectory generation, and stability analysis of an 8-degree-of-freedom (DOF) biped robot walking on flat and inclined terrain. Denavit--Hartenberg (DH) parameters and homogeneous transformations are used to derive the forward kinematics, while closed-form inverse kinematics maps the desired hip and swing-foot Cartesian trajectories, generated with cubic splines, to joint angles. Joint torques are computed with the Newton--Euler iterative algorithm, and dynamic stability is evaluated with the zero moment point (ZMP) criterion. A genetic algorithm (GA) optimizes the hip height, maximum swing-foot lift, and frontal-plane tilt angle by minimizing the work done by the joints subject to a ZMP feasibility penalty. Simulation results in MATLAB show that the nominal 8-DOF model remains ZMP-stable for step completion times down to 0.5~s and for slope inclinations up to 22.5$^\circ$ with the given foot geometry; beyond these limits the ZMP leaves the support polygon and either the foot dimensions or the trajectory parameters must be modified. The results also show that ZMP stability is governed by the mass distribution among the links rather than the total mass of the robot.
\end{abstract}


\section{Introduction}
Bipedal robots are of practical interest because their morphology matches environments built for humans, avoiding the need for a separate infrastructure~\cite{vukobratovic2012biped}. A central difficulty is that the kinematic chain of the robot changes between an open chain in the single support phase (SSP) and a closed chain in the double support phase (DSP), for which a unique solution generally does not exist~\cite{vukobratovic2012biped}. This work models, simulates, and stabilizes the walk of an 8-DOF planar/spatial biped on flat and inclined ground using the ZMP criterion, and reports the operating envelope (step time, slope angle) over which the nominal design remains dynamically stable.

\section{Related Work}
The ZMP criterion, introduced by Vukobratovi\'{c}~\cite{vuko}, remains the most widely used stability test for biped gait synthesis. Kajita \emph{et al.}~\cite{kajita_zmp} generated walking patterns by preview control of the ZMP using a cart-table model, improving on the discrete Fourier-based approach of Takanishi \emph{et al.}~\cite{hashimoto2015online}. Yang \emph{et al.}~\cite{yang2007uniform} combined GA optimization with truncated Fourier series for gait trajectories, while Shin \emph{et al.}~\cite{shin} and Zhu \emph{et al.}~\cite{zhu} used ZMP- and energy-based objectives, respectively, to reduce actuator energy consumption. Alternative stability measures include the foot-rotation indicator of Goswami~\cite{goswami1999postural}, the contact-wrench criterion of Hirukawa~\cite{adios_zmp}, and the foot-placement estimator of Wight \emph{et al.}~\cite{wight2008introduction}. Walking on uneven and inclined terrain has been addressed with trunk compensation~\cite{takanishi1990realization}, compliant control~\cite{hyon2008compliant}, terrain-sensing feet~\cite{kang2010realization}, and learning-based step planning~\cite{gupta2018trajectory}. The present work follows the ZMP/GA line of research and quantifies the flat- and inclined-ground stability envelope for a low-DOF, hardware-representative biped.

\section{Kinematic Model}
\subsection{DH Parameterization}
\begin{figure}[t]
\centering
\includegraphics[width=0.72\linewidth]{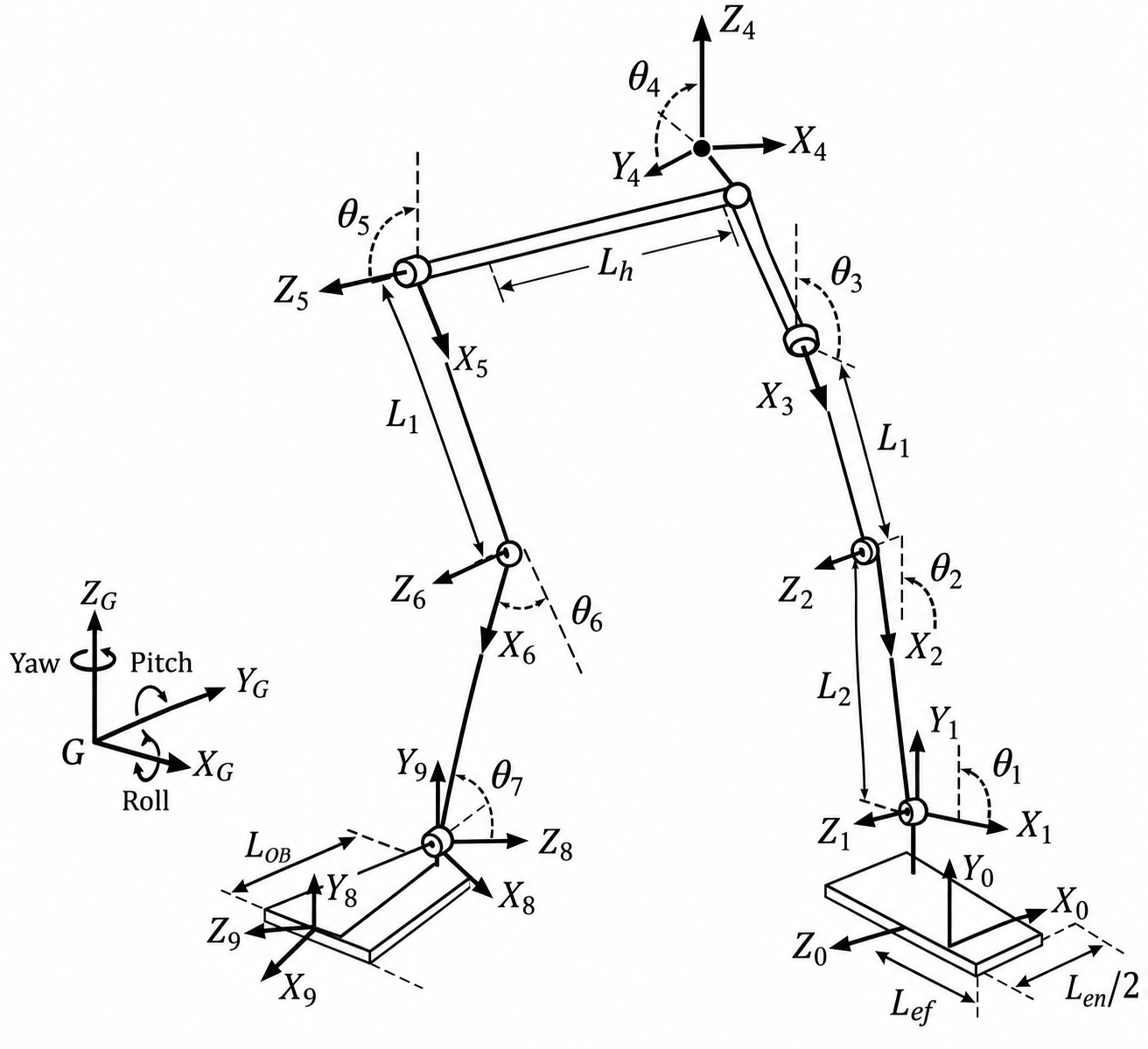}
\caption{Frame assignment for the 8-DOF biped robot.}
\label{fig:dh}
\end{figure}

The biped has 8~DOF: two at each ankle (sagittal and frontal tilt), one at each knee, and one at each hip, which is the minimum configuration required for full 3-D motion without kinematic redundancy. Frames are assigned to each joint following the Denavit--Hartenberg convention~\cite{lipkin2005note}, and consecutive frames are related by the homogeneous transform
\begin{equation}
{}^{i}T_{i+1}=
\begin{pmatrix}
c\theta & -s\theta & 0 & a\\
s\theta c\alpha & c\theta c\alpha & -s\alpha & -s\alpha\, d\\
s\theta s\alpha & c\theta s\alpha & c\alpha & c\alpha\, d\\
0 & 0 & 0 & 1
\end{pmatrix},
\label{eq:dh}
\end{equation}
where $c(\cdot)=\cos(\cdot)$, $s(\cdot)=\sin(\cdot)$, and $a,\alpha,d,\theta$ are the DH parameters~\cite{craig}. Table~\ref{tab:dh} lists the parameters for the stance-toe-to-swing-toe chain (Fig.~\ref{fig:dh}). Given the hip and swing-ankle trajectories, the full-body configuration follows from ${}^{1}T_{8}={}^{1}T_{2}\,{}^{2}T_{3}\cdots{}^{7}T_{8}$~\cite{fu1987robotics}.

\begin{table}[t]
\centering
\caption{DH parameters of the 8-DOF biped}
\label{tab:dh}
\begin{tabular}{|c|c|c|c|c|}
\hline
Link & $a$ & $\alpha$ & $d$ & $\theta$\\
\hline
0--1 & 0 & 0 & 0 & $\theta_1$\\
1--2 & 0 & $-\pi/2$ & 0 & $\theta_2$\\
2--3 & $l$ & 0 & 0 & $\theta_3$\\
3--4 & $l$ & 0 & 0 & $\theta_4$\\
4--5 & 0 & 0 & $l_h$ & $-(\pi/2+\theta_5)$\\
5--6 & $l$ & 0 & 0 & $-\theta_6$\\
6--7 & $l$ & 0 & 0 & $\theta_7$\\
7--8 & 0 & $\pi/2$ & 0 & $\theta_8$\\
\hline
\end{tabular}
\end{table}

\subsection{Walking on an Inclined Surface}
On a slope of angle $\theta_s$, only the ground-to-stance-toe transform changes, since the toe-to-ankle transform is terrain-independent~\cite{vundavilli2011balanced}:
\begin{equation}
T_{gf}=
\begin{pmatrix}
c\theta_s & -s\theta_s & 0 & \left(\tfrac{l_{step}}{2}+l_{af}\right)c\theta_s\\
0 & 0 & -1 & l_h/2\\
s\theta_s & c\theta_s & 0 & \left(\tfrac{l_{step}}{2}+l_{af}\right)s\theta_s\\
0 & 0 & 0 & 1
\end{pmatrix}\!,
\label{eq:incline}
\end{equation}
which is implemented by rotating the foot frame by the surface inclination~\cite{sugahara2005walking}. Walking is decomposed into SSP and DSP, occupying 60\% and 40\% of the cycle, respectively~\cite{mitobe1997control,kajita2014biped}.

\section{Trajectory Generation and Inverse Kinematics}
The hip and swing-foot Cartesian trajectories are fitted with MATLAB cubic splines through prescribed via points~\cite{sarkar20158,arakawa1996natural}: the hip moves $L_{step}/2$ in $x$ at constant height $H_z$, while the swing foot moves $L_{step}$ in $x$ and rises to a peak height $h_{max}$ in $z$. The frontal-plane side-tilt is likewise a cubic in the ankle angles, $\theta_1(t)=d_0+d_1t+d_2t^2+d_3t^3$ with $\theta_8=-\theta_1$. Given these Cartesian trajectories, closed-form inverse kinematics~\cite{shih1993inverse} yields the sagittal-plane joint angles:
\begin{align}
\psi&=\arctan\frac{O_{x4}-O_{x2}}{O_{z4}-O_{z2}},\\
d&=\sqrt{(O_{z4}-O_{z2})^2+(O_{x4}-O_{x2})^2},\\
\phi&=\arccos\frac{L_1^2+d^2-L_2^2}{2L_2 d}, \qquad
\theta_2=\psi-\phi,\\
\theta_3&=2\phi\;(L_1{=}L_2), \qquad \theta_4=\theta_3-\theta_2,
\end{align}
with the swing-leg angles $\theta_5,\theta_6,\theta_7$ obtained analogously.

\section{Dynamics and Stability Criterion}
\subsection{Newton--Euler Iterative Dynamics}
Joint torques are found with the Newton--Euler iterative method~\cite{orin1979kinematic,ardema2004newton}: an outward recursion computes each link's angular/linear velocity and acceleration, from which the inertial force ${}^{i+1}F_{i+1}=m_{i+1}{}^{i+1}\dot v_{c_{i+1}}$ and torque ${}^{i+1}N_{i+1}={}^{i+1}I_{c_{i+1}}{}^{i+1}\dot\Omega_{i+1}+{}^{i+1}\Omega_{i+1}\times{}^{i+1}I_{c_{i+1}}{}^{i+1}\Omega_{i+1}$ are obtained; an inward recursion then propagates joint forces/moments and the actuator torque $\tau_i={}^{i}n_i^{T}\,{}^{i}\hat z_i$.

\subsection{Zero Moment Point}
The ZMP is the point on the ground where the net moment of the ground reaction about the $x$ and $y$ axes vanishes~\cite{vukobratovic2001zero}. For $n$ point masses,
\begin{equation}
x_{zmp}=\frac{\sum_{i=1}^{n}\!\big[I_i\dot\omega_i+m_i x_i(\ddot z_i-g)-m_i\ddot x_i z_i\big]}{\sum_{i=1}^{n}m_i(\ddot z_i-g)},
\label{eq:zmp}
\end{equation}
with $y_{zmp}$ defined analogously~\cite{dekker2009zero}. Reducing the biped to a linear inverted pendulum model (LIPM) — constant center-of-mass (CoM) height, zero angular momentum about the CoM — simplifies \eqref{eq:zmp} to~\cite{kajita2003biped}
\begin{equation}
P_x = x_{com}+\frac{z_{com}\,\ddot x_{com}}{g}, \qquad
P_y = y_{com}+\frac{z_{com}\,\ddot y_{com}}{g}.
\label{eq:lipm}
\end{equation}
The gait is dynamically stable if $(x_{zmp},y_{zmp})$ remains within the convex hull of the support foot polygon at every instant; otherwise the robot tips over.

\section{Trajectory Optimization by Genetic Algorithm}
The via-point trajectories of Section~IV leave three free parameters — hip height $H_z$, maximum swing-foot lift $Sz_{max}$, and maximum frontal tilt $\theta_{1,max}$ — that are tuned with a GA~\cite{whitley1994genetic,mukhopadhyay2009genetic} because it requires no gradient information~\cite{choi1999optimal}. The objective combines the total work done by the eight joints with a ZMP feasibility penalty~\cite{sarkar2019optimal}:
\begin{equation}
O = w + p,\qquad
w=\sum_{n=1}^{8}\int \tau_n\,\dot\theta_n\,dt,\qquad
p = a\times10^{4},
\label{eq:ga}
\end{equation}
where $a=0$ if the ZMP lies inside the support polygon and $a=1$ otherwise. MATLAB's \texttt{ga} solver was run with a population of 130, 500 generations, a crossover fraction of 0.95, and a mutation rate of 0.05.

\section{Results and Discussion}
\subsection{Simulation Setup}
Link lengths and joint limits are taken from the Bioloid humanoid kit so the model is directly realizable in hardware. Key parameters: foot length 0.10~m, ankle-to-foot offset $L_{af}=0.049$~m, ankle height $L_{an}=0.032$~m, lower-trunk (hip) width $L_h=0.085$~m, and thigh/shank length $L_1{=}L_2=0.111$~m; step length equals $L_1$ and the nominal swing-foot lift is $0.6\,L_1$. Walking is simulated in MATLAB.

\subsection{Effect of Step Completion Time on Flat Ground}
\begin{figure}[t]
\centering
\includegraphics[width=0.85\linewidth]{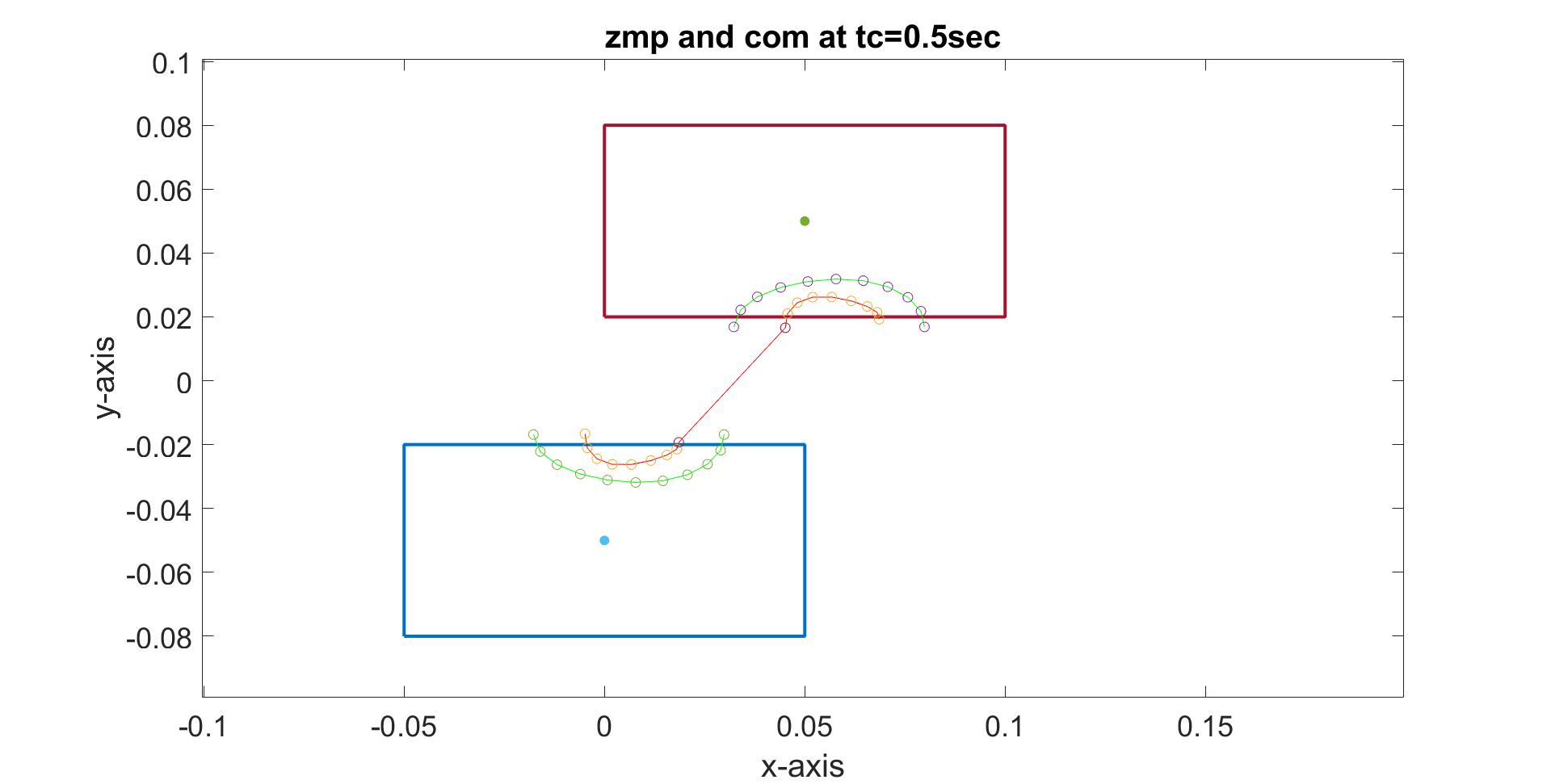}
\caption{ZMP (orange) and CoM (green) trajectories on flat ground for a step completion time of $t_c=0.5$~s; the rectangle is the support-foot polygon.}
\label{fig:zmp05}
\end{figure}

For a step time $t_c=2$~s the ZMP and CoM traces nearly coincide (Fig.~\ref{fig:zmp05}, left case) and the gait is comfortably stable. As $t_c$ is reduced, the ZMP trace shifts progressively toward the boundary of the foot polygon; at $t_c=0.5$~s (Fig.~\ref{fig:zmp05}) it approaches the polygon edge, and at $t_c=0.4$~s it moves almost entirely outside, i.e., the given foot area cannot sustain a step faster than about 0.5~s without redesign.

\subsection{Walking on an Inclined Surface}
\begin{figure}[t]
\centering
\includegraphics[width=0.85\linewidth]{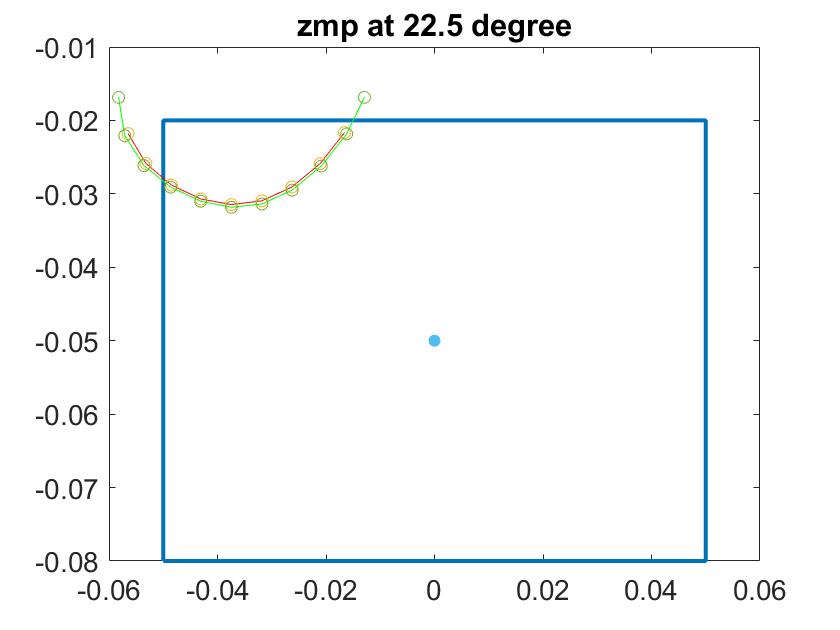}
\caption{ZMP trajectory at $22.5^\circ$ incline: the trace has shifted toward the rear of the support polygon relative to flat-ground walking.}
\label{fig:incline}
\end{figure}

Applying the same trajectory on a slope shifts the mass backward. The ZMP remains fully inside the foot polygon up to $15^\circ$; at $18^\circ$ and $22.5^\circ$ it shifts increasingly toward the rear edge (Fig.~\ref{fig:incline}), and beyond $22.5^\circ$ it leaves the polygon under the nominal foot dimensions. Re-optimizing the trajectory with the GA for the $22.5^\circ$ case gives $\theta_{1,max}{=}{-9.18^\circ}$, $H_z{=}0.151$~m, $Sz_{max}{=}0.013$~m, restoring a ZMP trace fully inside the polygon. These parameters do not transfer to $30^\circ$: the ZMP violates the foot length even after re-optimization (best fitness saturates at the penalty value), and stability is recovered only after the foot length is extended by 0.01~m (from 0.05 to 0.06~m), after which GA optimization again yields a feasible trajectory. Also note that friction angles of most surfaces (0.4--0.6) may be insufficient to prevent slip near this incline, which further motivates keeping the slope limit at $22.5^\circ$ for the nominal design.

\subsection{Joint Torques and GA-Optimized Trajectories}
\begin{figure}[t]
\centering
\includegraphics[width=0.95\linewidth]{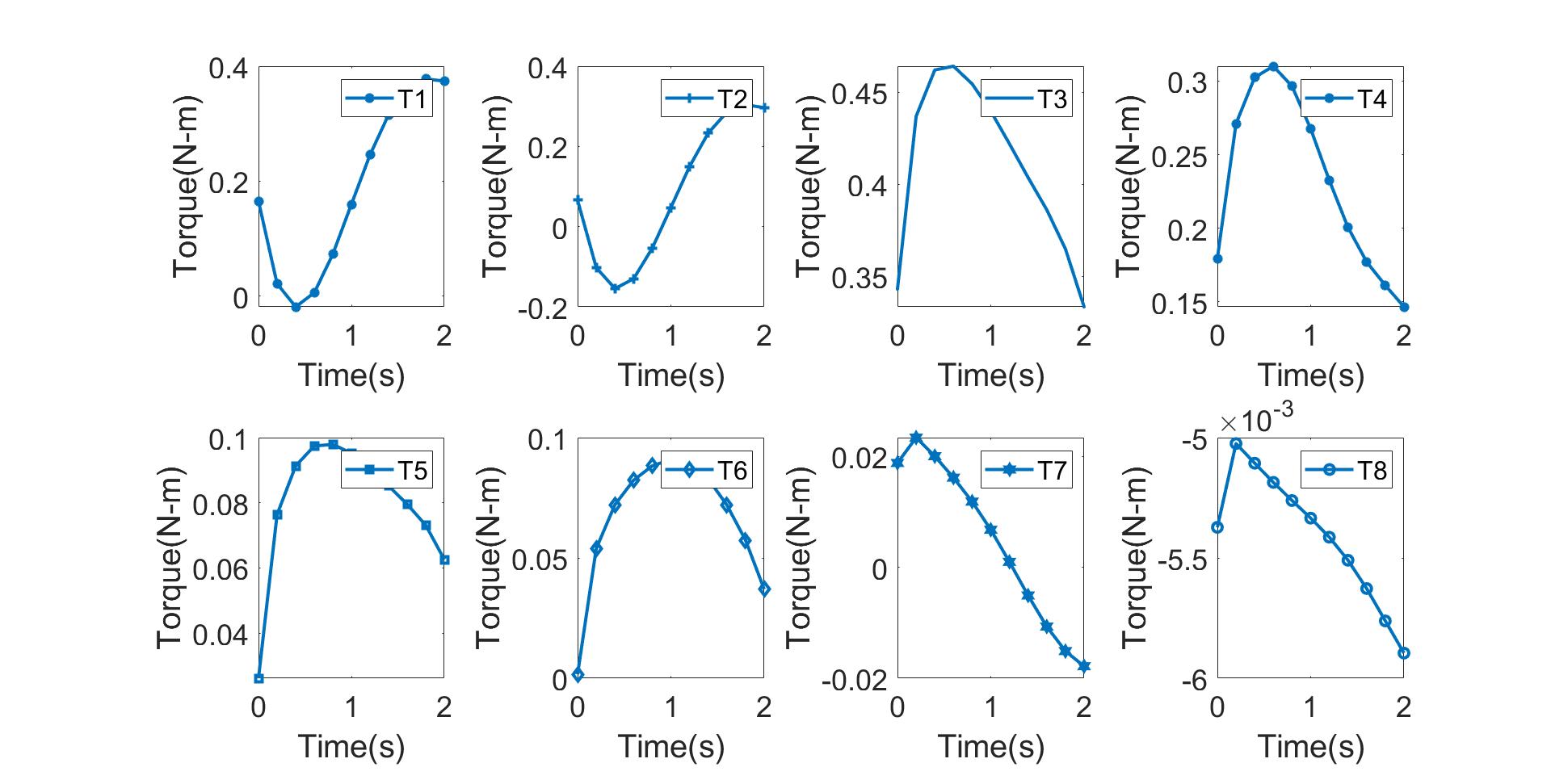}
\caption{Newton--Euler joint torques during flat-ground walking. Stance-leg joints carry more torque than swing-leg joints.}
\label{fig:torque}
\end{figure}

\begin{figure}[t]
\centering
\includegraphics[width=0.85\linewidth]{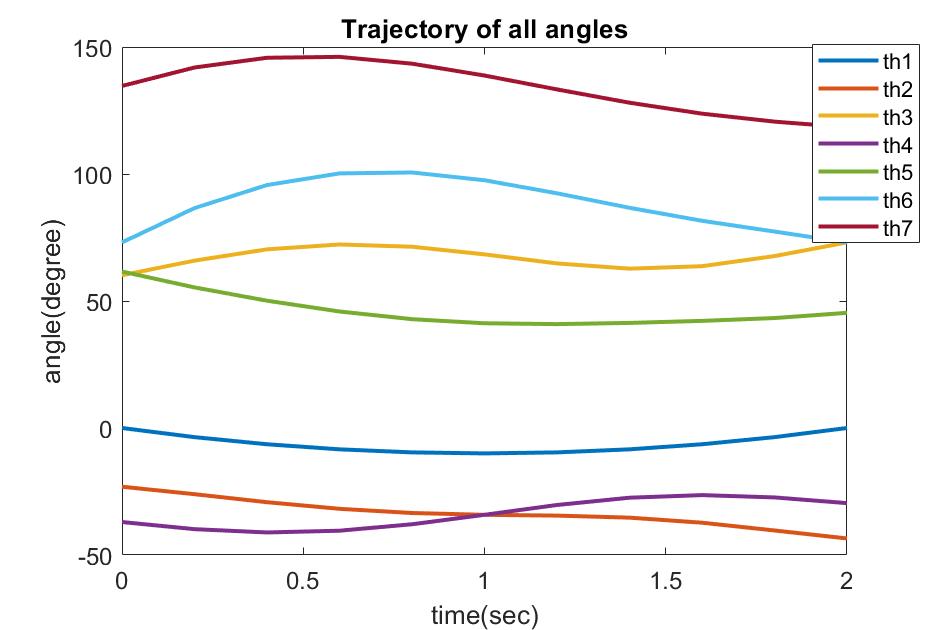}
\caption{GA-optimized joint-angle trajectories $\theta_1$--$\theta_7$ over one step.}
\label{fig:joints}
\end{figure}

Fig.~\ref{fig:torque} shows the Newton--Euler torque profile at each joint during flat-ground walking. Stance-leg joints require substantially higher torque than swing-leg joints because they must react the dynamic load transmitted through the entire kinematic chain; each joint torque rises and then decreases smoothly over the step, consistent with the cubic-spline trajectory. Fig.~\ref{fig:joints} shows the resulting GA-optimized joint-angle trajectories $\theta_1$--$\theta_7$ used to produce these torques.

\subsection{Effect of Mass Distribution and Comparison with the LIPM}
Scaling the mass of every link by 10$\times$ left the ZMP trace of the full 8-DOF model unchanged, but doubling only the swing-leg mass pushed the ZMP outside the polygon, i.e., ZMP stability depends on how mass is distributed across the links, not on the total mass. Comparing the full 8-DOF model with the LIPM of \eqref{eq:lipm} across step times from 2~s down to 0.5~s shows near-identical ZMP traces at low acceleration; at higher acceleration (shorter $t_c$) the two diverge, with the 8-DOF model becoming unstable while the LIPM still predicts stability, confirming that the LIPM's constant-height, zero-angular-momentum assumptions break down under fast, high-acceleration steps.

\section{Conclusion and Future Work}
An 8-DOF biped was modeled with DH-based forward/inverse kinematics, cubic-spline trajectory generation, Newton--Euler dynamics, and ZMP-based stability checking, with the hip height, swing-foot lift, and frontal tilt tuned by a work-minimizing GA under a ZMP feasibility constraint. On the nominal (Bioloid-scale) foot geometry, the simulated gait is stable for step times down to 0.5~s on flat ground and for slopes up to $22.5^\circ$; beyond these limits either the trajectory or the foot dimensions must be modified, and ZMP stability was shown to depend on link-wise mass distribution rather than total mass. As this is a non-redundant, fully actuated model, control is comparatively simple; future work will extend the approach to stepping over obstacles and to higher-DOF bipeds where kinematic redundancy must be resolved online.

\bibliographystyle{IEEEtran}
\bibliography{refs}

@article{vuko,
  title={Zero-moment point—thirty five years of its life},
  author={Vukobratovi{\'c}, Miomir and Borovac, Branislav},
  journal={International journal of humanoid robotics},
  volume={1},
  number={01},
  pages={157--173},
  year={2004},
  publisher={World Scientific}
}

@INPROCEEDINGS{kajita_zmp,  author={Kajita, S. and Kanehiro, F. and Kaneko, K. and Fujiwara, K. and Harada, K. and Yokoi, K. and Hirukawa, H.},  booktitle={2003 IEEE International Conference on Robotics and Automation (Cat. No.03CH37422)},   title={Biped walking pattern generation by using preview control of zero-moment point},   year={2003},  volume={2},  number={},  pages={1620-1626 vol.2},  doi={10.1109/ROBOT.2003.1241826}}

@book{craig,
  title={Introduction to robotics: mechanics and control, 3/E},
  author={Craig, John J},
  year={2009},
  publisher={Pearson Education India}
}

@incollection{hashimoto2015online,
  title={Online walking pattern generation using FFT for humanoid robots},
  author={Hashimoto, Kenji and Kondo, Hideki and Lim, Hun-Ok and Takanishi, Atsuo},
  booktitle={Motion and Operation Planning of Robotic Systems},
  pages={417--438},
  year={2015},
  publisher={Springer}
}

@article{yang2007uniform,
  title={A uniform biped gait generator with offline optimization and online adjustable parameters},
  author={Yang, Lin and Chew, Chee-Meng and Zielinska, Teresa and Poo, Aun-Neow},
  journal={Robotica},
  volume={25},
  number={5},
  pages={549},
  year={2007},
  publisher={Cambridge University Press}
}

@ARTICLE{shin,  author={Shin, Hyeok-Ki and Kim, Byung Kook},  journal={IEEE Transactions on Robotics},   title={Energy-Efficient Gait Planning and Control for Biped Robots Utilizing the Allowable ZMP Region},   year={2014},  volume={30},  number={4},  pages={986-993},  doi={10.1109/TRO.2014.2305792}}

@article{zhu,
  title={Energy-efficient bio-inspired gait planning and control for biped robot based on human locomotion analysis},
  author={Zhu, Hongbo and Luo, Minzhou and Mei, Tao and Zhao, Jianghai and Li, Tao and Guo, Fayong},
  journal={Journal of Bionic Engineering},
  volume={13},
  number={2},
  pages={271--282},
  year={2016},
  publisher={Springer}
}

@article{goswami1999postural,
  title={Postural stability of biped robots and the foot-rotation indicator (FRI) point},
  author={Goswami, Ambarish},
  journal={The International Journal of Robotics Research},
  volume={18},
  number={6},
  pages={523--533},
  year={1999},
  publisher={SAGE Publications}
}

@article{wight2008introduction,
  title={Introduction of the foot placement estimator: A dynamic measure of balance for bipedal robotics},
  author={Wight, Derek L and Kubica, Eric G and Wang, David WL},
  journal={Journal of computational and nonlinear dynamics},
  volume={3},
  number={1},
  year={2008},
  publisher={American Society of Mechanical Engineers Digital Collection}
}

@inproceedings{adios_zmp,
  title={A universal stability criterion of the foot contact of legged robots-adios zmp},
  author={Hirukawa, Hirohisa and Hattori, Shizuko and Harada, Kensuke and Kajita, Shuuji and Kaneko, Kenji and Kanehiro, Fumio and Fujiwara, Kiyoshi and Morisawa, Mitsuharu},
  booktitle={Proceedings 2006 IEEE International Conference on Robotics and Automation, 2006. ICRA 2006.},
  pages={1976--1983},
  year={2006},
  organization={IEEE}
}

@inproceedings{takanishi1990realization,
  title={Realization of dynamic biped walking stabilized by trunk motion on a sagittally uneven surface},
  author={Takanishi, Atsuo and Lim, Hun-ok and Tsuda, Masayuki and Kato, Ichiro},
  booktitle={EEE International Workshop on Intelligent Robots and Systems, Towards a New Frontier of Applications},
  pages={323--330},
  year={1990},
  organization={IEEE}
}

@article{hyon2008compliant,
  title={Compliant terrain adaptation for biped humanoids without measuring ground surface and contact forces},
  author={Hyon, Sang-Ho},
  journal={IEEE Transactions on Robotics},
  volume={25},
  number={1},
  pages={171--178},
  year={2008},
  publisher={IEEE}
}

@inproceedings{kang2010realization,
  title={Realization of biped walking on uneven terrain by new foot mechanism capable of detecting ground surface},
  author={Kang, Hyun-jin and Hashimoto, Kenji and Kondo, Hideki and Hattori, Kentaro and Nishikawa, Kosuke and Hama, Yuichiro and Lim, Hun-ok and Takanishi, Atsuo and Suga, Keisuke and Kato, Keisuke},
  booktitle={2010 IEEE International Conference on Robotics and Automation},
  pages={5167--5172},
  year={2010},
  organization={IEEE}
}

@article{gupta2018trajectory,
  title={Trajectory generation and step planning of a 12 DoF biped robot on uneven surface},
  author={Gupta, Gaurav and Dutta, Ashish},
  journal={Robotica},
  volume={36},
  number={7},
  pages={945--970},
  year={2018},
  publisher={Cambridge University Press}
}

@inproceedings{choi1999optimal,
  title={Optimal walking trajectory generation for a biped robot using genetic algorithm},
  author={Choi, Sang-Ho and Choi, Young-Ha and Kim, Jin-Geol},
  booktitle={Proceedings 1999 IEEE/RSJ International Conference on Intelligent Robots and Systems. Human and Environment Friendly Robots with High Intelligence and Emotional Quotients (Cat. No. 99CH36289)},
  volume={3},
  pages={1456--1461},
  year={1999},
  organization={IEEE}
}

@article{sarkar2019optimal,
  title={Optimal trajectory generation and design of an 8-dof compliant biped robot for walk on inclined ground},
  author={Sarkar, Abhishek and Dutta, Ashish},
  journal={Journal of Intelligent \& Robotic Systems},
  volume={94},
  number={3},
  pages={583--602},
  year={2019},
  publisher={Springer}
}

@article{whitley1994genetic,
  title={A genetic algorithm tutorial},
  author={Whitley, Darrell},
  journal={Statistics and computing},
  volume={4},
  number={2},
  pages={65--85},
  year={1994},
  publisher={Springer}
}

@article{mukhopadhyay2009genetic,
  title={Genetic algorithm: A tutorial review},
  author={Mukhopadhyay, Deep Malya and Balitanas, Maricel O and Farkhod, Alisherov and Jeon, Seung-Hwan and Bhattacharyya, Debnath},
  journal={International journal of grid and distributed computing},
  volume={2},
  number={3},
  pages={25--32},
  year={2009}
}

@inproceedings{vukobratovic2001zero,
  title={Zero moment point-Proper interpretation and new applications},
  author={Vukobratovi{\'c}, Miomir and Borovac, Branislav and {\v{S}}urdilovi{\'c}, Dragoljub},
  booktitle={Int. Conf. on Humanoid Robots},
  pages={237--244},
  year={2001}
}

@article{dekker2009zero,
  title={Zero-moment point method for stable biped walking},
  author={Dekker, MHP},
  journal={Eindhoven University of Technology},
  year={2009}
}

@inproceedings{kajita2003biped,
  title={Biped walking pattern generation by using preview control of zero-moment point},
  author={Kajita, Shuuji and Kanehiro, Fumio and Kaneko, Kenji and Fujiwara, Kiyoshi and Harada, Kensuke and Yokoi, Kazuhito and Hirukawa, Hirohisa},
  booktitle={2003 IEEE International Conference on Robotics and Automation (Cat. No. 03CH37422)},
  volume={2},
  pages={1620--1626},
  year={2003},
  organization={IEEE}
}

@book{ardema2004newton,
  title={Newton-Euler Dynamics},
  author={Ardema, Mark D},
  year={2004},
  publisher={Springer Science \& Business Media}
}

@article{orin1979kinematic,
  title={Kinematic and kinetic analysis of open-chain linkages utilizing Newton-Euler methods},
  author={Orin, David E and McGhee, RB and Vukobratovi{\'c}, M and Hartoch, G},
  journal={Mathematical Biosciences},
  volume={43},
  number={1-2},
  pages={107--130},
  year={1979},
  publisher={Elsevier}
}

@article{shih1993inverse,
  title={Inverse kinematics and inverse dynamics for control of a biped walking machine},
  author={Shih, Ching-Long and Gruver, William A and Lee, Tsu-Tian},
  journal={Journal of Robotic Systems},
  volume={10},
  number={4},
  pages={531--555},
  year={1993},
  publisher={Wiley Online Library}
}

@inproceedings{arakawa1996natural,
  title={Natural motion trajectory generation of biped locomotion robot using genetic algorithm through energy optimization},
  author={Arakawa, Takemasa and Fukuda, Toshio},
  booktitle={1996 IEEE International Conference on Systems, Man and Cybernetics. Information Intelligence and Systems (Cat. No. 96CH35929)},
  volume={2},
  pages={1495--1500},
  year={1996},
  organization={IEEE}
}

@article{sarkar20158,
  title={8-DoF biped robot with compliant-links},
  author={Sarkar, Abhishek and Dutta, Ashish},
  journal={Robotics and Autonomous Systems},
  volume={63},
  pages={57--67},
  year={2015},
  publisher={Elsevier}
}

@article{mitobe1997control,
  title={Control of a biped walking robot during the double support phase},
  author={Mitobe, Kazuhisa and Mori, Naoki and Nasu, Yasuo and Adachi, N},
  journal={Autonomous Robots},
  volume={4},
  number={3},
  pages={287--296},
  year={1997},
  publisher={Springer}
}

@article{vundavilli2011balanced,
  title={Balanced gait generations of a two-legged robot on sloping surface},
  author={Vundavilli, Pandu Ranga and Pratihar, Dilip Kumar},
  journal={Sadhana},
  volume={36},
  number={4},
  pages={525},
  year={2011},
  publisher={Springer}
}

@inproceedings{sugahara2005walking,
  title={Walking control method of biped locomotors on inclined plane},
  author={Sugahara, Yusuke and Mikuriya, Yutaka and Hashimoto, Kenji and Hosobata, Takuya and Sunazuka, Hiroyuki and Kawase, Masamiki and Lim, Hun-ok and Takanishi, Atsuo},
  booktitle={Proceedings of the 2005 IEEE International Conference on Robotics and Automation},
  pages={1977--1982},
  year={2005},
  organization={IEEE}
}

@book{fu1987robotics,
  title={Robotics: Control Sensing. Vis.},
  author={Fu, King Sun and Gonzalez, Ralph and Lee, CS George},
  year={1987},
  publisher={Tata McGraw-Hill Education}
}

@inproceedings{lipkin2005note,
  title={A note on Denavit-Hartenberg notation in robotics},
  author={Lipkin, Harvey},
  booktitle={International Design Engineering Technical Conferences and Computers and Information in Engineering Conference},
  volume={47446},
  pages={921--926},
  year={2005}
}

@incollection{kajita2014biped,
  title={Biped walking},
  author={Kajita, Shuuji and Hirukawa, Hirohisa and Harada, Kensuke and Yokoi, Kazuhito},
  booktitle={Introduction to humanoid robotics},
  pages={105--158},
  year={2014},
  publisher={Springer}
}

@book{vukobratovic2012biped,
  title={Biped locomotion: dynamics, stability, control and application},
  author={Vukobratovic, Miomir and Borovac, Branislav and Surla, Dusan and Stokic, Dragan},
  volume={7},
  year={2012},
  publisher={Springer Science \& Business Media}
}

\end{document}